%% file: ms.tex
\documentclass[a4paper,twoside]{article}

\input preamble.tex

\usepackage{multicol}
\usepackage{pslatex}
\usepackage{apalike}

\usepackage{SCITEPRESS}

\begin{document}

\title{Transfer Learning for Image-Based Malware Classification}

\author{\authorname{Niket Bhodia\sup{1}, Pratikkumar Prajapati\sup{1}, Fabio Di Troia\sup{1} and Mark Stamp\sup{1}}
\affiliation{\sup{1}Department of Computer Science, San Jose State University, San Jose, California, USA}
\email{niket.bhodia@sjsu.edu, pratikkumar.prajapati@sjsu.edu, fabioditroia@msn.com, mark.stamp@sjsu.edu}
}


\abstract{In this paper, we consider the problem of malware detection and classification based on image analysis.
We convert executable files to images and apply image recognition using deep learning (DL) models. 
To train these models, we employ transfer learning based on existing DL models that have been pre-trained 
on massive image datasets. We carry out various experiments with this technique and compare its 
performance to that of an extremely simple machine learning technique, namely, $k$-nearest neighbors (\kNN).
For our $k$-NN experiments, we use features extracted directly from executables, rather than image analysis. 
While our image-based DL technique performs well in the experiments, 
surprisingly, it is outperformed by~\kNN. We show that DL models are 
better able to generalize the data,  in the sense that they outperform~\kNN\ in
simulated zero-day experiments.}

\onecolumn \maketitle \normalsize \setcounter{footnote}{0} \vfill

\section{\uppercase{Introduction}}\label{sect:intro}

\noindent Traditionally, malware detection has relied on 
pattern matching against signatures extracted from known malware. While simple 
and efficient, signature scanning is easily defeated by a number of well-known evasive strategies. 
This fact has given rise to statistical and machine learning based detection techniques, 
which are more robust to code modification. In response, malware writers have developed
advanced forms of malware that alter statistical and 
structural properties of their code. Such ``noise'' 
can cause statistical models to misclassify samples.

In this paper, we compare image-based deep learning (DL) models for malware analysis
to a much simpler non-image based technique. To train these DL models, we employ transfer learning,
relying on models that have been pre-trained on large image datasets. Leveraging the power 
of such models has been shown to yield strong malware detection and classification results~\cite{PB2}.
Intuitively, we might expect that models based on image analysis to be more robust,
as compared to models that rely on opcodes, byte $n$-grams,
or similar statistical features~\cite{DamodaranTVAS17},
\cite{SinghTVAS16},
\cite{TodericiS13},
\cite{BaysaLS13},
\cite{AustinFJS13},
\cite{WongS06}.

To the best of our knowledge, image analysis was 
first applied to the malware problem in~\cite{PB1}, where
high-level ``gist'' descriptors were used. 
More recently,~\cite{PB2} confirmed these results and contrasted the gist-descriptor method to 
a DL approach that produced equally good---if not slightly better---results without the extra 
work required to extract gist descriptors.
A direct comparison to more straightforward machine learning techniques 
seems to be lacking in previous work, making it difficult to determine the 
comparative advantages and disadvantages of DL image-based analysis in the 
malware domain.

In this paper, we extend the analysis found in~\cite{PB2} in various directions.
For example, we consider improvements to the DL training,
and we apply our improved image-based DL approach to a more challenging 
dataset. Most significantly, we compare the performance of image-based DL analysis 
to a relatively simple and straightforward non-image based strategy using
$k$-nearest neighbors (\kNN). These~\kNN\ experiments
yield somewhat surprising results and serve to highlight the strengths
and weaknesses of DL image-based analysis.

\section{\uppercase{Methodology}}\label{sect:meth}

\noindent In this section, we discuss the datasets, data pre-processing, and
features extracted. We also discuss implementation details.

\subsection{Datasets}\label{sect:2.1}

We consider two malware datasets, namely,
Malimg~\cite{PB1} and Malicia~\cite{PB3}.
The Malimg dataset contains~9,339 malware images from~25 families, 
while Malicia has~11,668 malware binaries from~54 families.

The Malimg dataset consists of images, and hence these samples 
require no pre-processing before applying image-based analysis. 
However, the binaries corresponding to the Malimg images are not readily available.
In contrast, the Malicia samples are binaries and hence they
must be converted into images before
we can apply image-based analysis. We found that~581 samples from the Malicia dataset 
were not {\tt exe} files, and~1,192 samples did not have a family label.
These samples were excluded, leaving us with~9,895 binaries from~51 families
from the Malicia dataset.
 
The family breakdown for the Malimg and Malicia datasets are given in Tables~\ref{tab:malimg} 
and~\ref{tab:malicia}, respectively. In Table~\ref{tab:malimg}, we abbreviate ``password stealing'' as ``pws,''
``downloader'' as ``dl,'' and ``backdoor'' as ``bd.'' In Table~\ref{tab:malicia}, the ``other'' category
consists of~38 families, each of which has less than~10 samples per family, with the majority  
of these ``families'' contributing only a single sample.

\begin{table}[htb]
\caption{Malimg dataset}\label{tab:malimg}
\centering
\begin{tabular}{llr}
\hline\hline
		{Family} & {Type} & {Samples} \\ \toprule
		Adialer.C &  dialer     & 122 \\
		Agent.FYI &   bd    & 116 \\
		Allaple.A &  worm     & 2,949 \\
		Allaple.L &  worm     & 1,591 \\
		Alueron.gen!J &  trojan     & 198 \\
		Autorun.K &  worm     & 106 \\
		C2LOP.gen!g & trojan     &  200 \\
		C2LOP.P &  trojan     & 146 \\
		Dialplatform.B &  dialer     & 177 \\
		Dontovo.A &  dl     & 162 \\
		Fakerean &   rogue    & 381 \\
		Instantaccess &  dialer     & 431 \\
		Lolyda.AA1 &  pws     & 213 \\
		Lolyda.AA2 &  pws     & 184 \\
		Lolyda.AA3 &  pws     & 123 \\
		Lolyda.AT &   pws    & 159 \\
		Malex.gen!J &   trojan    & 136 \\
		Obfuscator.AD & dl      & 142 \\
		Rbot!gen &   bd    & 158 \\
		Skintrim.N &    trojan    & 80 \\
		Swizzor.gen!E &   dl    & 128 \\
		Swizzor.gen!I &  dl     & 132 \\
		VB.AT &  worm     & 408 \\
		Wintrim.BX &   dl     & 97 \\
		Yuner.A &  worm     & 800 \\ 
\hline
		Total &  ---   &  9,339 \\
\hline\hline
\end{tabular}
\end{table}

\begin{table}[htb]
\caption{Malicia dataset}\label{tab:malicia}
\centering
\begin{tabular}{lrc}
\hline\hline
		{Family} & {Samples} & {Size} \\ \toprule
		cleaman &         32 & small \\
		CLUSTER:46.105.131.121 &         20 & small \\
		CLUSTER:85.93.17.123 &         45 & small \\
		CLUSTER:astaror &         24 & small \\
		CLUSTER:newavr &         29  & small \\
		CLUSTER:positivtkn.in.ua &         14  & small \\
		cridex &         74  & small \\
		harebot & 53  & small \\
		securityshield & 150 & large \\
		smarthdd &         68 & small  \\
		winwebsec &       5,820  & large \\
		zbot &       2,167 & large \\
		zeroaccess &       1,306 & large \\
		other (38 families) & 93  & small \\
\hline
		Total &      9,895  & --- \\
\hline\hline
\end{tabular}
\end{table}

In addition, two benign datasets were used. The first of these benign sets consists
of~3304 binaries typically found on a modern Windows PC. Our second
benign dataset contains~704 binaries from the Cygwin library. 

\subsection{Data Preprocessing}\label{sect:2.2}

Our DL method requires images as input. For Malimg, we directly 
use the images that comprise the dataset---the only preprocessing 
involves separating the images into training and validation sets. 
For Malicia, we have malware binaries, which are converted to images 
by adapting the script used by the authors of~\cite{PB1}.
More details on this image conversion process
are provided in Section~\ref{sect:2.3}. 

For our \kNN\ experiments, we do not use images, but instead extract a set of features
directly from binaries. More details on these features are provided in Section~\ref{sect:2.4}. 
Since we did not have access to Malimg binaries, we could not test our \kNN\ approach
on this dataset. 
We compare our \kNN\ results to image-based
DL using the the Malicia samples.

The Malicia dataset is highly unbalanced---four families 
dominate, as can be seen from the counts
in Table~\ref{tab:malicia}. Hence, we have partitioned the dataset 
into two parts, with one set containing only samples from the large families 
and one containing all samples from the small families, where
we consider any family with more than~100 samples to be ``large.''
Both of these Malicia subsets are used in different variations of our experiments.


\subsection{Converting Binaries to Images}\label{sect:2.3}

To convert a binary to an image we treat the sequence of bytes representing
the binary as the bytes of a grayscale PNG image. In all of our experiments, we
use a predefined width of~256, and a variable length, depending on the size of the binary.

\begin{figure}[!htb]
	\centering
	\includegraphics[width=0.45\textwidth]{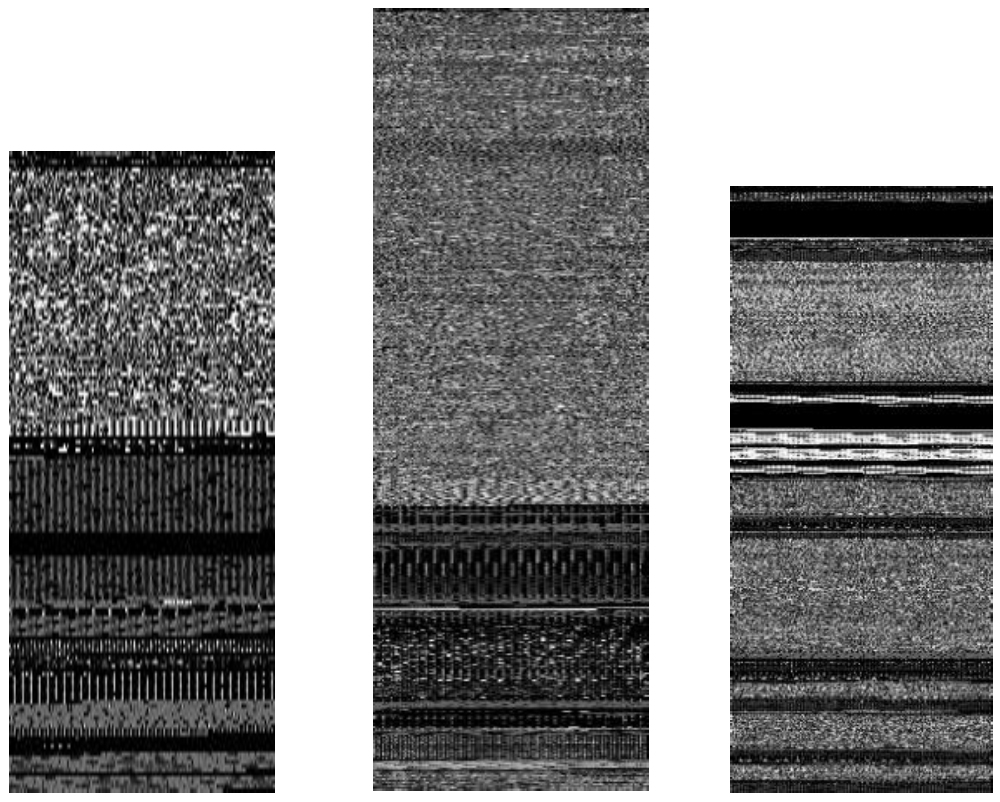}
	\caption{Unrelated binaries as images}\label{fig:fig1}
\end{figure}

\begin{figure}[!htb]
	\centering
	\includegraphics[width=0.45\textwidth]{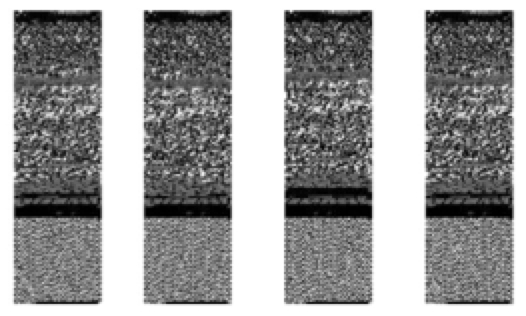}
	\caption{Variants of malware from the Malimg family of \hbox{Dialplatform.B} 
	as images~\cite{PB3}}\label{fig:figVars}
\end{figure}

Sample images of unrelated binaries are given in Figure~\ref{fig:fig1},
while samples from a malware family appear in Figure~\ref{fig:figVars}.
From these examples, the allure of image-based
classification is clear---images tend to smooth out minor
within-family differences, while significant (i.e., between family) 
differences are clearly observed.

\subsection{Feature Extraction for \kNN}\label{sect:2.4}

We adapted code from two publicly accessible GitHub repositories~\cite{PB4}
and~\cite{PB5} 
to extract~54 features from each binary sample. For the sake
of brevity, we list~15 of these~54 features in Table~\ref{tab:kNN_features},
where feature names are listed in the left-hand column,
while the right-hand column gives the feature value extracted from the
benign sample~{\rm VC\underline{\phantom{\rm ii}}redist.x64}.

\begin{table}[!htb]
\caption{Examples of \kNN\ features}\label{tab:kNN_features}
\centering
\footnotesize
\begin{tabular}{lr} \hline\hline
Name & {\rm VC\underline{\phantom{\rm ii}}redist.x64} \\ \toprule
%
SizeOfOptionalHeader & 224 \\
SizeOfCode & 234496 \\
%
FileAlignment & 512 \\
MajorOSVersion & 5 \\
%
SizeOfImage & 413696 \\
SizeOfHeaders & 1024 \\
Subsystem & 2 \\
SizeOfStackCommit & 4096 \\
%
SectionsNb & 7 \\
SectionsMeanEntropy & 3.7137 \\
SectionMaxRawsize & 234496 \\
%
SectionMaxVirtualsize & 234372 \\
ImportsNb & 285 \\
ResourcesMaxEntropy & 5.2550 \\
%
ResourcesMaxSize & 9652 \\
\hline\hline
\end{tabular}
\end{table}

\subsection{Implementation Details}\label{sect:2.5}

The DL models were implemented using the {\rm fast.ai} library~\cite{PB10}, 
which is built on top of the {\rm PyTorch} framework. The choice of this library was 
influenced by the fact that it incorporates several DL best practices,
including learning rate finding, stochastic gradient descent with restarts, 
and differential learning rates. 

For \kNN, we used the popular {\rm Scikit-learn} library~\cite{PB7}, which is based on 
many of the fundamentals described in~\cite{StampML2017}.
The {\rm fast.ai} library incorporates CUDA support, which allowed us to
accelerate the training process by making use of the graphics card.

\section{\uppercase{Experiments and Results}}\label{sect:exp}

\noindent We performed a variety of experiments involving various combinations of datasets, 
classification level (binary and multiclass), and learning techniques (DL and \kNN).
Here, we present results for eight separate experiments, 
as listed in Table~\ref{tab:exp}. Each experiment represented a 
specific combination of datasets, classification level, and learning technique. 
In the remainder of this section, we discuss each of these experiments in some detail.

\begin{table*}[htb]
\caption{Experiments}\label{tab:exp}
\centering
\begin{tabular}{clllll} 
\hline\hline
\multirow{2}{*}{Number} & \multirow{2}{*}{Classification} & Malware & Benign & Learning & \multirow{2}{*}{Accuracy} \\
                                       &                                                & dataset   & dataset & technique \\ \toprule
1    & binary & Malimg & Windows & DL & 98.39\%\ \\ 
2    & multiclass (26) & Malimg & Windows & DL & 94.80\%\ \\ 
3    & binary & Malicia (large) & Windows & DL & 97.61\%\ \\ 
4    & multiclass (5) & Malicia (large) & Windows & DL & 92.93\%\ \\ 
5    & binary & Malicia (large) & Windows & \kNN & 99.60\%\ \\ 
6    & multiclass (5) & Malicia (large) & Windows & \kNN & 99.43\%\ \\ 
7   & binary (zero-day) & Malicia (small) & Cygwin &  DL & 91.17\%\ \\ 
8   & binary (zero-day) & Malicia (small) & Cygwin &  \kNN & 89.00\%\ \\ 
\hline\hline
\end{tabular}
\end{table*}


For the DL experiments, that is, experiments~1 through~4 in Table~\ref{tab:exp}, we 
tested variants of the ResNet model~\cite{PB9}, specifically, ResNet34, ResNet50, ResNet101, 
and ResNext50. We chose ResNet because of its combination of performance and efficiency. 
ResNet-based architectures won the ImageNet and COCO challenges in~2015. Their key 
advantage is the use of ``residual blocks,'' which enabled the training of neural networks
of unprecedented depth. The models we use were pre-trained on the ImageNet dataset, 
which contains some~1.2 million images in~1,000 classes. 

The more complex ResNet variants we experimented with 
did not yield significant improvement, so we used ResNet34 for all DL experiments reported in this paper. 
We also tested various combinations of hyperparameters, including the number of epochs, the learning rate, 
the number of cycles of learning rate annealing, and variations in the cycle length. 
The training concepts implemented in conjunction with these hyperparameters were 
cosine annealing, learning rate finding, stochastic gradient descent with restarts, 
freezing and unfreezing layers in the pre-trained network, and differential learning rates. 
A description of these techniques and how they are used in concert with the listed 
hyperparameters is beyond the scope of this paper---the interested reader can refer 
to~\cite{PB2},~\cite{PB10}, and~\cite{PB11} for more details.

Perhaps the simplest machine learning technique possible 
is~\kNN, where we classify a sample
based on its~$k$ nearest neighbors in a given training set. For~\kNN, there
is no explicit training phase, and all work is deferred to the scoring phase.
Once the training data is specified, we score a sample by simply
determining its nearest neighbors in the training set, with a majority
vote typically used for (binary) classification.
In spite of its incredible simplicity, it is often the case
that \kNN\ achieves results that are competitive
with far more complex machine learning techniques~\cite{StampML2017}. 

For our \kNN\ experiments (i.e., experiments~5 and~6), we use Euclidean distance, and hence
the only parameter to be determined is the value of~$k$, that is,
the number of neighbors to consider when classifying a sample. We experimented
with values ranging from~$k=1$ to~$k=9$, and we found that the best results were obtained 
with~$k = 1$, as can be seen in both Figures~\ref{fig:exp5}(a) and~\ref{fig:exp6}(a). 
Thus, we have used~$k=1$ for the \kNN\ results presented in this paper.
Again, for these experiments, the feature vector consists of~54
PE file features extracted using modified forms of the code at~\cite{PB4}
and~\cite{PB5}.

\section{\uppercase{Discussion}}\label{sect:disc}

\noindent For our first set of experiments, we apply the image-based
DL technique outlined above to the Malimg dataset. 
We consider the following two variations.

\begin{description}
\item[Experiment~1] For our first experiment, we perform binary classification 
of malware versus benign, where the malware class is obtained 
by simply grouping all Malimg families into one malware set. 
The benign set consists of~3304 Windows samples, which have been converted to images.
\item[Experiment~2] For the corresponding multiclass classification problem, 
we attempt to classify the malware samples into their 
respective families, with the Windows benign set treated as an additional ``family.'' 
Since there are~25 malware families in the Malimg dataset, for this classification problem, 
we have~26 classes.
\end{description}

For the binary classification problem in experiment~1, we obtained an accuracy of~98.39\%,
while the multiclass problem in experiment~2 yielded an accuracy of~94.80\%.
The results of experiment~1 are summarized in Figure~\ref{fig:exp1}, 
while Figures~\ref{fig:exp2a} and~\ref{fig:exp2b} give the results for experiment~2. 
These experimental results are comparable to 
those obtained in~\cite{PB2}, and serve to confirm our DL implementation.

\begin{figure*}[!htb]
	\centering
	\begin{tabular}{cc}
	\input figures/train1.tex
	&
	\input figures/conf_exp1b.tex
	\\
	(a) Training
	&
	(b) Confusion matrix
	\\[2ex]
	\end{tabular}
	\caption{Experiment~1 results}\label{fig:exp1}
\end{figure*}

\begin{figure}[!htb]
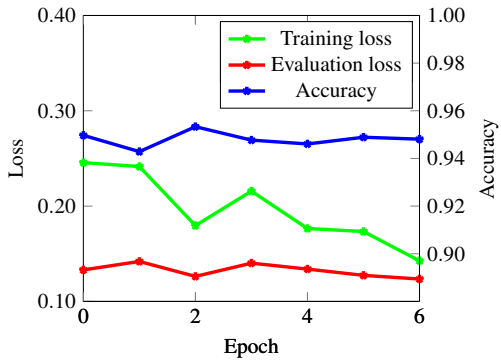

	\centering
	\input figures/train2.tex
	\caption{Experiment~2 training}\label{fig:exp2a}
\end{figure}

\begin{figure}[!htb]
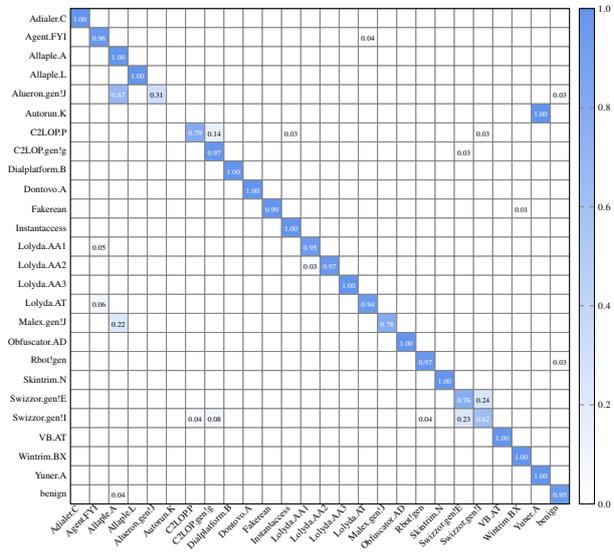

	\centering
	\input figures/conf_exp2.tex
	\caption{Experiment~2 confusion matrix}\label{fig:exp2b}
\end{figure}

We do not have access to the Malimg binary files, so we are unable to compare the DL results 
for this dataset to alternatives that rely on features extracted directly from executables.
Therefore, we next consider the Malicia malware dataset, which will allow us to 
compare our image-based DL technique to a simpler \kNN\ analysis based on
non-image features.

\begin{figure*}[!htb]
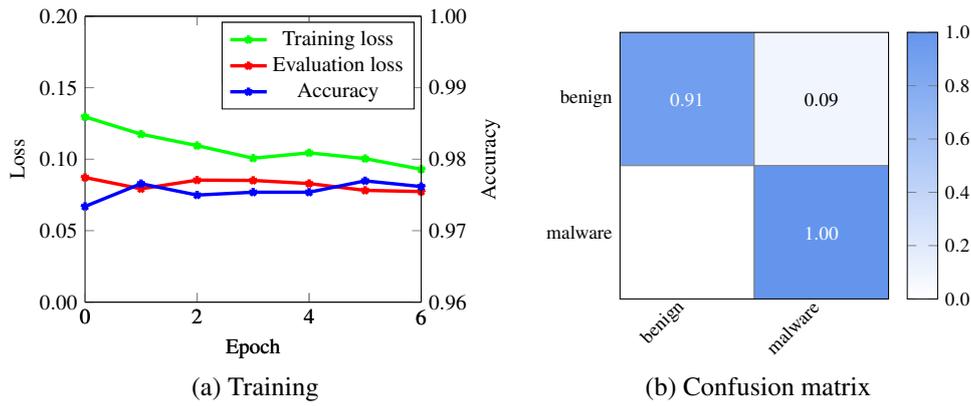

	\centering
	\begin{tabular}{cc}
	\input figures/train3.tex
	&
	\input figures/conf_exp3a.tex
	\\
	(a) Training
	&
	(b) Confusion matrix
	\\[2ex]
	\end{tabular}
	\caption{Experiment~3}\label{fig:exp3}
\end{figure*}

\begin{figure*}[!htb]
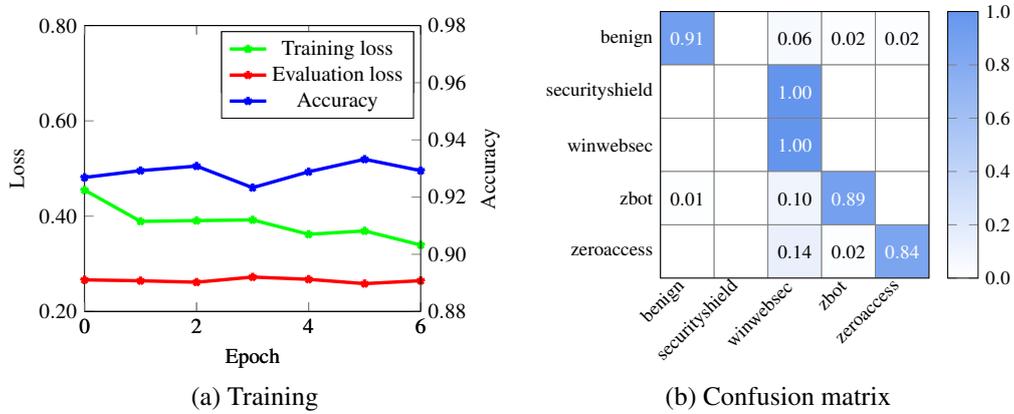

	\centering
	\begin{tabular}{cc}
	\input figures/train4.tex
	&
	\input figures/conf_exp4.tex
	\\
	(a) Training
	&
	(b) Confusion matrix
	\\[2ex]
	\end{tabular}
	\caption{Experiment~4}\label{fig:exp4}
\end{figure*}

For the Malicia dataset, we first generate an image corresponding to each binary
executable sample in the dataset, as discussed in Section~\ref{sect:2.3}.
Then we perform the analogous experiments to~1 and~2, above, but using 
the Malicia samples in place of Malimg. Specifically, we perform the following experiments.
\begin{description}
\item[Experiment~3] As in experiment~1, we perform binary classification of 
malware versus benign, but in this case, the malware class consists of all 
Malicia samples, as images. The benign set consists of the same~3304 Windows
samples that were used in experiment~1.
\item[Experiment~4] For the corresponding multiclass version of this problem,
we attempt to classify the Malicia (image) samples into their 
respective families, with the Windows benign set treated as an additional ``family.'' 
\end{description}
For the binary classification problem in experiment~3, we obtain an
accuracy of~97.61\%, while the multiclass problem in experiment~4 yields
a classification accuracy of~92.93\%. The results of experiment~3
are summarized in Figure~\ref{fig:exp3}, while Figure~\ref{fig:exp4}
contains the results of experiment~4. Note that only the four large Malicia families were used in these 
experiments, as the remaining families are severely underrepresented in the dataset.
These results indicate that the multiclass problem
is far more challenging for the Malicia dataset, as compared to the
Malimg dataset. Recall that there are~26 classes in the
Malimg classification experiment, but only five classes in the corresponding Malicia
experiment, yet we obtain a lower multiclass accuracy on the Malicia samples.

Next, we compare our DL approach to a simpler strategy
based on \kNN. We extract non-image features from the 
Malicia binaries and the benign set, as discussed in Section~\ref{sect:2.4}.
Then we carry out binary and multiclass experiments.
Specifically, we perform the following \kNN\ experiments.
\begin{description}
\item[Experiment~5] For this binary classification experiment, we deal with malware and benign sets, 
where the malware class consists of Malicia samples. 
In this case, non-image features are extracted directly from the malware binaries.
The benign set again consists of the~3304 Windows samples, 
and the same non-image features have been extracted from these samples.
\item[Experiment~6] In the corresponding multiclass experiment, 
we attempt to categorize the Malicia samples into their 
respective families, with the Windows benign set treated as a
yet another ``family.'' As above, here we only use the four large 
Malicia families which, together with the benign set, gives us a total
of five distinct classes.
\end{description}
As mentioned above, we selected \kNN\ for these experiments because we 
want to establish a baseline by which to compare the performance of
our image-based DL approach. We also want to use non-image features in this alternative
analysis, as this provides some additional insight into the value of treating malware samples as images. 

Interestingly, \kNN\ outperforms DL, achieving an impressively high accuracy 
of~99.60\%\ in the binary classification problem, while a similarly high accuracy of~99.43\%\ is
attained in the multiclass problem. Figures~\ref{fig:exp5} and~\ref{fig:exp6}, respectively,
summarize the results of experiment~5 and experiment~6. Note that
the multiclass result in experiment~6 is particularly strong, given that there are five
classes under consideration, including a benign set. In contrast, our image-based DL
technique yielded substantially worse results, 
with an accuracy of less than~93\%\ on this same dataset.

\begin{figure*}[!htb]
	\centering
	\begin{tabular}{ccc}
	\input figures/optK_5.tex
	&
	&
	\input figures/conf_exp5a_b.tex \\
	(a) Determining optimal~$k$
	&
	&
	(b) Confusion matrix
	\\[2ex]
	\end{tabular}
	\caption{Experiment~5}\label{fig:exp5}
\end{figure*}


\begin{figure*}[!htb]
	\centering
	\begin{tabular}{ccc}
	\input figures/optK_6.tex
	&
	&
	\input figures/conf_exp6.tex \\
	(a) Determining optimal~$k$
	&
	&
	(b) Confusion matrix
	\\[2ex]
	\end{tabular}
	\caption{Experiment~6}\label{fig:exp6}
\end{figure*}


Next, we attempt to quantify the robustness and generalizability of our DL (image-based) technique
in comparison to our \kNN\ ({\tt exe}-based) classification strategy. For the DL and \kNN\ cases, 
denoted here as experiments~7 and~8, respectively, we attempt to classify samples as malware
or benign, based on samples belonging 
to families that the models have not been trained to detect.
This can be viewed as simulating zero-day malware, that is, 
malware that was not available during the training phase. 
Specifically, we performed the following zero-day experiments.

\begin{figure*}[!htb]
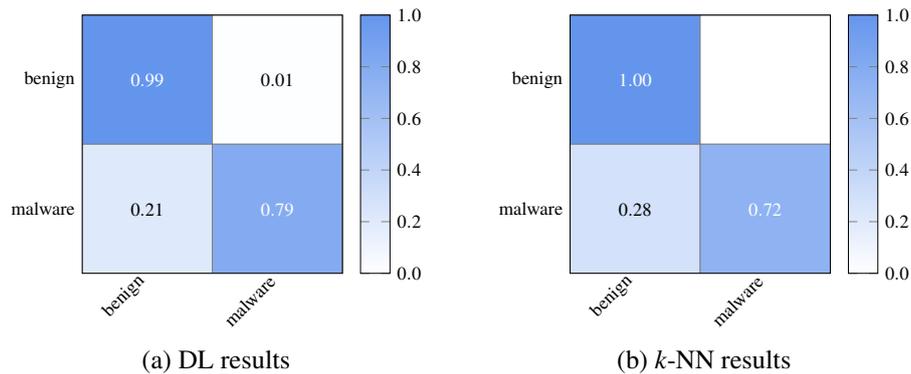

	\centering
	\input figures/conf_zd.tex
	\caption{Zero-day simulations (experiments~7 and~8)}\label{fig:exp78}
\end{figure*}

\begin{description}
\item[Experiment~7] We test our DL approach for the binary classification 
of zero-day malware versus benign, where the malware training
set consists of all samples in the four large Malicia families. The benign training set consists
of~3304 Windows samples. To simulate zero-day malware, the test set consists of 
all of the small families 
in the Malicia dataset. In addition, to ensure that unfamiliar benign binaries did not 
lead to a high false positive rate, we used~704 Cygwin binaries as our benign test set.  
\item[Experiment~8] For our corresponding \kNN\ experiments, we use the
same datasets as in experiment~7. 
And, as above, to simulate zero-day malware, 
the malware test set consists of all of the small families in the Malicia dataset, and the benign test set
consists of the~704 Cygwin samples. 
\end{description}

Our image-based DL model performed reasonably well in this zero-day simulation, 
correctly identifying~79\%\ of the malware 
samples, with a low false positive rate of~1\%.
However, our DL model has a high false
negative rate, as illustrated in Figure~\ref{fig:exp78}~(a). 
With~\kNN, we achieve broadly similar, but somewhat worse results,
as can be seen from the confusion matrix in Figure~\ref{fig:exp78}.
These zero-day experiments indicate that
image-based DL models generalize somewhat better than a more
straightforward~\kNN\ model.
This is a potentially an advantage for image-based DL models
in the malware realm, as detecting zero-day
malware is the holy grail in the AV field. However, the simplicity 
and ease of training~\kNN\ models could be a major advantage
in some situations.

\section{\uppercase{Conclusion}}\label{sect:con}

\noindent In this paper, 
we treated malware binaries as images and classified samples 
based on pre-trained deep learning image recognition models. 
We compared these image-based deep learning (DL)
results to a simpler $k$-nearest neighbor (\kNN) approach 
based on a more typical set of static features.
We carried out a wide variety of experiments, each representing a 
different combination of dataset, 
classification level, 
and learning technique. 
The multiclass experiments were particularly impressive, with high accuracy 
attained over a large number of malware families.

Our DL method overall delivered results comparable to previous work, yet
it was outperformed by the much simpler~\kNN\ learning technique in
some cases. The image-based DL models 
did outperform~\kNN\ in simulated zero-day experiments, which
indicates that this DL implementation better generalizes the training 
data, as compared to~\kNN. This is a significant point, since zero-day malware,
arguably, represents the ultimate challenge in malware detection.

There are many promising avenues for future work related to image-based malware analysis. 
For example, it seems likely that a major strength of any image-based strategy is 
its robustness. Consequently, additional experiments along these lines
would be helpful to better quantify this effect. 

\bibliographystyle{apalike}
{\small
\bibliography{other.bib,Stamp-Mark.bib}}

\end{document}

%% file: preamble.tex
\usepackage{amsmath,amsthm, amsfonts, amssymb, amsxtra,amsopn}
\usepackage{pgfplots}
\pgfplotsset{compat=1.13}
\usepackage{pgfplotstable}
\usepackage{graphicx}
\usepackage{multirow}
\usepackage{tabu}
\usepackage{algorithmic}
\usepackage{longtable}
\usepackage{booktabs}
\usepackage{listings}
\usepackage{cmap}
\usepackage{colortbl}
\usepackage{adjustbox}
\usepackage{epsfig}

\usepackage{subfigure}

\PassOptionsToPackage{hyphens}{url}

\usepackage{hyperref}
\hypersetup{colorlinks=true,linkcolor=black,citecolor=black,urlcolor=blue,filecolor=black}
\hypersetup{pdfpagemode=UseNone,pdfstartview=}

\usepackage[tableposition=top]{caption}
\usepackage{enumitem}
\setlist[itemize]{noitemsep, topsep=0pt}

\usepackage{SCITEPRESS}     

\def\kNN{{$k$-NN}}

\pgfkeys{
    /pgf/number format/fixed zerofill=true }
    
\pgfplotstableset{
    color cells/.code={%
        \pgfqkeys{/color cells}{#1}%
        \pgfkeysalso{%
            postproc cell content/.code={%
                \begingroup
                \pgfkeysgetvalue{/pgfplots/table/@preprocessed cell content}\value
\ifx\value\empty
\endgroup
\else
                \pgfmathfloatparsenumber{\value}%
                \pgfmathfloattofixed{\pgfmathresult}%
                \let\value=\pgfmathresult
                \pgfplotscolormapaccess
                    [\pgfkeysvalueof{/color cells/min}:\pgfkeysvalueof{/color cells/max}]%
                    {\value}%
                    {\pgfkeysvalueof{/pgfplots/colormap name}}%
                \pgfkeysgetvalue{/pgfplots/table/@cell content}\typesetvalue
                \pgfkeysgetvalue{/color cells/textcolor}\textcolorvalue
                \toks0=\expandafter{\typesetvalue}%
                \xdef\temp{%
                    \noexpand\pgfkeysalso{%
                        @cell content={%
                            \noexpand\cellcolor[rgb]{\pgfmathresult}%
                            \noexpand\definecolor{mapped color}{rgb}{\pgfmathresult}%
                            \ifx\textcolorvalue\empty
                            \else
                                \noexpand\color{\textcolorvalue}%
                            \fi
                            \the\toks0 %
                        }%
                    }%
                }%
                \endgroup
                \temp
\fi
            }%
        }%
    }
}

%% file: figures/train1.tex
\begin{tikzpicture}[scale=0.8]
\begin{axis}[axis y line*=left,
		   width=0.45\textwidth,
		   height=0.4\textwidth,
		   /pgf/number format/1000 sep={},
                    xmin=0,xmax=6,
                    ymin=0.0,ymax=0.2,
                    xticklabel style={
                    	/pgf/number format/fixed,
			/pgf/number format/precision=0},
                    yticklabel style={
                    	/pgf/number format/fixed,
			/pgf/number format/precision=2},
                    xlabel={Epoch},
                    ylabel={Loss}] 
\addplot[color=green,ultra thick,mark=star,mark size=2.0] coordinates {
(0,0.075045)
(1,0.074993)
(2,0.058729)
(3,0.059499)
(4,0.063076)
(5,0.052170)
(6,0.048897)
}; \label{plot_one1}
\addlegendentry{Training loss}
\addplot[color=red,ultra thick,mark=star,mark size=2.0] coordinates {
(0,0.044364)
(1,0.043749)
(2,0.040201)
(3,0.049734)
(4,0.040368)
(5,0.039356)
(6,0.037642)
}; \label{plot_two1}
\addlegendentry{Evaluation loss}
\end{axis}
\begin{axis}[axis y line*=right,
		   width=0.45\textwidth,
		   height=0.4\textwidth,
		   /pgf/number format/1000 sep={},
                    xmin=0,xmax=6,
                    ymin=0.96,ymax=1.0,
                    xticklabel style={
                    	/pgf/number format/fixed,
			/pgf/number format/precision=0},
                    yticklabel style={
                    	/pgf/number format/fixed,
			/pgf/number format/precision=2},
                    xlabel={Epoch},
                    ylabel={Accuracy}] 
\addlegendimage{/pgfplots/refstyle=plot_one1}\addlegendentry{Training loss}
\addlegendimage{/pgfplots/refstyle=plot_two1}\addlegendentry{Evaluation loss}
\addplot[color=blue,ultra thick,mark=star,mark size=2.0] coordinates {
(0,0.980282)
(1,0.981489)
(2,0.981489)
(3,0.980684)
(4,0.983903)
(5,0.982294)
(6,0.983903)
}; \label{plot_three1}
\addlegendentry{Accuracy}
\end{axis}
\end{tikzpicture}

%% file: figures/conf_exp1b.tex
\begin{tikzpicture}[scale=0.8]
    \begin{axis}[
        width=6cm,
        height=6cm,
	colormap={bluewhite}{color=(white) rgb255=(100,149,237)},
        xticklabels={benign,malware},
        xtick={0,...,1},
        xtick style={draw=none},
	xticklabel style={anchor=east,rotate=45},
        yticklabels={benign,malware},
        ytick={0,...,1},
        ytick style={draw=none},
        enlargelimits=false,
        xlabel style={font=\footnotesize},
        ylabel style={font=\footnotesize},
        legend style={font=\footnotesize},
        xticklabel style={font=\footnotesize},
        yticklabel style={font=\footnotesize},
        colorbar,
        colorbar style={
            ytick={0,0.20,0.40,0.60,0.80,1.00},
            yticklabels={0,0.20,0.40,0.60,0.80,1.00},
            yticklabel={\pgfmathprintnumber\tick},
            yticklabel style={font=\footnotesize,
            		/pgf/number format/fixed,
			/pgf/number format/precision=1}
        },
        point meta min=0.0,
        point meta max=1.0,
        nodes near coords={\pgfmathprintnumber\pgfplotspointmeta},
        nodes near coords black white/.style={
            small value/.style={
                font=\footnotesize,
                yshift=-7pt,
                text=black,
                /pgf/number format/fixed,
                /pgf/number format/precision=3
            },
            large value/.style={
                font=\footnotesize,
                yshift=-7pt,
                text=white,
                /pgf/number format/fixed,
                /pgf/number format/precision=3
            },
            every node near coord/.style={
                check for zero/.code={
                    \pgfmathfloatifflags{\pgfplotspointmeta}{0}{
                        \pgfkeys{/tikz/coordinate}
                    }{
                        \begingroup
                        \pgfkeys{/pgf/fpu}
                        \pgfmathparse{\pgfplotspointmeta<#1}
                        \global\let\result=\pgfmathresult
                        \endgroup
                        %
                        %
                        \pgfmathfloatcreate{1}{1.0}{0}
                        \let\ONE=\pgfmathresult
                        \ifx\result\ONE
                            \pgfkeysalso{/pgfplots/small value}
                        \else
                            \pgfkeysalso{/pgfplots/large value}
                        \fi
                    }
                },
                check for zero,
            },
        },
        nodes near coords black white=0.5,
    ]
        \addplot[
            matrix plot,
            mesh/cols=2,
            point meta=explicit,draw=gray
        ] table [meta=C] {
            x y C
            0 0 0.960
            1 0 0.040
            0 1 0.002
            1 1 0.998
        };
    \end{axis}
\end{tikzpicture}
%

%% file: figures/train2.tex
\begin{tikzpicture}[scale=0.8]
\begin{axis}[axis y line*=left,
		   width=0.45\textwidth,
		   height=0.4\textwidth,
		   /pgf/number format/1000 sep={},
                    xmin=0,xmax=6,
                    ymin=0.1,ymax=0.4,
                    xticklabel style={
                    	/pgf/number format/fixed,
			/pgf/number format/precision=0},
                    yticklabel style={
                    	/pgf/number format/fixed,
			/pgf/number format/precision=2},
                    xlabel={Epoch},
                    ylabel={Loss}] 
\addplot[color=green,ultra thick,mark=star,mark size=2.0] coordinates {
(0,0.245607)
(1,0.241501)
(2,0.179555)
(3,0.215675)
(4,0.176497)
(5,0.173313)
(6,0.142592)
}; \label{plot_one2}
\addlegendentry{Training loss}
\addplot[color=red,ultra thick,mark=star,mark size=2.0] coordinates {
(0,0.133049)
(1,0.141914)
(2,0.126216)
(3,0.140143)
(4,0.133896)
(5,0.127316)
(6,0.123475)
}; \label{plot_two2}
\addlegendentry{Evaluation loss}
\end{axis}
\begin{axis}[axis y line*=right,
		   width=0.45\textwidth,
		   height=0.4\textwidth,
		   /pgf/number format/1000 sep={},
                    xmin=0,xmax=6,
                    ymin=0.88,ymax=1.0,
		   ytick={0.90,0.92,0.94,0.96,0.98,1.00},
                    xticklabel style={
                    	/pgf/number format/fixed,
			/pgf/number format/precision=0},
                    yticklabel style={
                    	/pgf/number format/fixed,
			/pgf/number format/precision=2},
                    xlabel={Epoch},
                    ylabel={Accuracy}] 
\addlegendimage{/pgfplots/refstyle=plot_one2}\addlegendentry{Training loss}
\addlegendimage{/pgfplots/refstyle=plot_two2}\addlegendentry{Evaluation loss}
\addplot[color=blue,ultra thick,mark=star,mark size=2.0] coordinates {
(0,0.949698)
(1,0.942857)
(2,0.953320)
(3,0.947686)
(4,0.946076)
(5,0.948893)
(6,0.948089)
}; \label{plot_three2}
\addlegendentry{Accuracy}
\end{axis}
\end{tikzpicture}

%% file: figures/conf_exp2.tex
\begin{tikzpicture}[scale=0.4]
    \begin{axis}[
        width=18cm,
        height=18cm,
	colormap={bluewhite}{color=(white) rgb255=(100,149,237)},
        xticklabels={Adialer.C,Agent.FYI,Allaple.A,Allaple.L,Alueron.gen!J,Autorun.K,C2LOP.P,C2LOP.gen!g,Dialplatform.B,Dontovo.A,Fakerean,Instantaccess,Lolyda.AA1,Lolyda.AA2,Lolyda.AA3,Lolyda.AT,Malex.gen!J,Obfuscator.AD,Rbot!gen,Skintrim.N,Swizzor.gen!E,Swizzor.gen!I,VB.AT,Wintrim.BX,Yuner.A,benign},
        xtick={0,...,25},
        xtick style={draw=none},
	xticklabel style={anchor=east,rotate=45},
        yticklabels={Adialer.C,Agent.FYI,Allaple.A,Allaple.L,Alueron.gen!J,Autorun.K,C2LOP.P,C2LOP.gen!g,Dialplatform.B,Dontovo.A,Fakerean,Instantaccess,Lolyda.AA1,Lolyda.AA2,Lolyda.AA3,Lolyda.AT,Malex.gen!J,Obfuscator.AD,Rbot!gen,Skintrim.N,Swizzor.gen!E,Swizzor.gen!I,VB.AT,Wintrim.BX,Yuner.A,benign},
        ytick={0,...,25},
        ytick style={draw=none},
        enlargelimits=false,
        xlabel style={font=\footnotesize},
        ylabel style={font=\footnotesize},
        legend style={font=\footnotesize},
        xticklabel style={font=\footnotesize},
        yticklabel style={font=\footnotesize},
        colorbar,
        colorbar style={
            ytick={0,0.20,0.40,0.60,0.80,1.00},
            yticklabels={0,0.20,0.40,0.60,0.80,1.00},
            yticklabel={\pgfmathprintnumber\tick},
            yticklabel style={font=\footnotesize,
            		/pgf/number format/fixed,
			/pgf/number format/precision=1}
        },
        point meta min=0.0,
        point meta max=1.0,
        nodes near coords={\pgfmathprintnumber\pgfplotspointmeta},
        nodes near coords black white/.style={
            small value/.style={
                font=\scriptsize,
                yshift=-7pt,
                text=black,
                /pgf/number format/fixed,
                /pgf/number format/precision=2
            },
            large value/.style={
                font=\scriptsize,
                yshift=-7pt,
                text=white,
                /pgf/number format/fixed,
                /pgf/number format/precision=2
            },
            every node near coord/.style={
                check for zero/.code={
                    \pgfmathfloatifflags{\pgfplotspointmeta}{0}{
                        \pgfkeys{/tikz/coordinate}
                    }{
                        \begingroup
                        \pgfkeys{/pgf/fpu}
                        \pgfmathparse{\pgfplotspointmeta<#1}
                        \global\let\result=\pgfmathresult
                        \endgroup
                        %
                        %
                        \pgfmathfloatcreate{1}{1.0}{0}
                        \let\ONE=\pgfmathresult
                        \ifx\result\ONE
                            \pgfkeysalso{/pgfplots/small value}
                        \else
                            \pgfkeysalso{/pgfplots/large value}
                        \fi
                    }
                },
                check for zero,
            },
        },
        nodes near coords black white=0.5,
    ]
        \addplot[
            matrix plot,
            mesh/cols=26,
            point meta=explicit,draw=gray
        ] table [meta=C] {
            x y C
0 0 1.00
1 0 0.00
2 0 0.00
3 0 0.00
4 0 0.00
5 0 0.00
6 0 0.00
7 0 0.00
8 0 0.00
9 0 0.00
10 0 0.00
11 0 0.00
12 0 0.00
13 0 0.00
14 0 0.00
15 0 0.00
16 0 0.00
17 0 0.00
18 0 0.00
19 0 0.00
20 0 0.00
21 0 0.00
22 0 0.00
23 0 0.00
24 0 0.00
25 0 0.00
0 1 0.00
1 1 0.96
2 1 0.00
3 1 0.00
4 1 0.00
5 1 0.00
6 1 0.00
7 1 0.00
8 1 0.00
9 1 0.00
10 1 0.00
11 1 0.00
12 1 0.00
13 1 0.00
14 1 0.00
15 1 0.04
16 1 0.00
17 1 0.00
18 1 0.00
19 1 0.00
20 1 0.00
21 1 0.00
22 1 0.00
23 1 0.00
24 1 0.00
25 1 0.00
0 2 0.00
1 2 0.00
2 2 1.00
3 2 0.00
4 2 0.00
5 2 0.00
6 2 0.00
7 2 0.00
8 2 0.00
9 2 0.00
10 2 0.00
11 2 0.00
12 2 0.00
13 2 0.00
14 2 0.00
15 2 0.00
16 2 0.00
17 2 0.00
18 2 0.00
19 2 0.00
20 2 0.00
21 2 0.00
22 2 0.00
23 2 0.00
24 2 0.00
25 2 0.00
0 3 0.00
1 3 0.00
2 3 0.00
3 3 1.00
4 3 0.00
5 3 0.00
6 3 0.00
7 3 0.00
8 3 0.00
9 3 0.00
10 3 0.00
11 3 0.00
12 3 0.00
13 3 0.00
14 3 0.00
15 3 0.00
16 3 0.00
17 3 0.00
18 3 0.00
19 3 0.00
20 3 0.00
21 3 0.00
22 3 0.00
23 3 0.00
24 3 0.00
25 3 0.00
0 4 0.00
1 4 0.00
2 4 0.67
3 4 0.00
4 4 0.31
5 4 0.00
6 4 0.00
7 4 0.00
8 4 0.00
9 4 0.00
10 4 0.00
11 4 0.00
12 4 0.00
13 4 0.00
14 4 0.00
15 4 0.00
16 4 0.00
17 4 0.00
18 4 0.00
19 4 0.00
20 4 0.00
21 4 0.00
22 4 0.00
23 4 0.00
24 4 0.00
25 4 0.03
0 5 0.00
1 5 0.00
2 5 0.00
3 5 0.00
4 5 0.00
5 5 0.00
6 5 0.00
7 5 0.00
8 5 0.00
9 5 0.00
10 5 0.00
11 5 0.00
12 5 0.00
13 5 0.00
14 5 0.00
15 5 0.00
16 5 0.00
17 5 0.00
18 5 0.00
19 5 0.00
20 5 0.00
21 5 0.00
22 5 0.00
23 5 0.00
24 5 1.00
25 5 0.00
0 6 0.00
1 6 0.00
2 6 0.00
3 6 0.00
4 6 0.00
5 6 0.00
6 6 0.79
7 6 0.14
8 6 0.00
9 6 0.00
10 6 0.00
11 6 0.03
12 6 0.00
13 6 0.00
14 6 0.00
15 6 0.00
16 6 0.00
17 6 0.00
18 6 0.00
19 6 0.00
20 6 0.00
21 6 0.03
22 6 0.00
23 6 0.00
24 6 0.00
25 6 0.00
0 7 0.00
1 7 0.00
2 7 0.00
3 7 0.00
4 7 0.00
5 7 0.00
6 7 0.00
7 7 0.97
8 7 0.00
9 7 0.00
10 7 0.00
11 7 0.00
12 7 0.00
13 7 0.00
14 7 0.00
15 7 0.00
16 7 0.00
17 7 0.00
18 7 0.00
19 7 0.00
20 7 0.03
21 7 0.00
22 7 0.00
23 7 0.00
24 7 0.00
25 7 0.00
0 8 0.00
1 8 0.00
2 8 0.00
3 8 0.00
4 8 0.00
5 8 0.00
6 8 0.00
7 8 0.00
8 8 1.00
9 8 0.00
10 8 0.00
11 8 0.00
12 8 0.00
13 8 0.00
14 8 0.00
15 8 0.00
16 8 0.00
17 8 0.00
18 8 0.00
19 8 0.00
20 8 0.00
21 8 0.00
22 8 0.00
23 8 0.00
24 8 0.00
25 8 0.00
0 9 0.00
1 9 0.00
2 9 0.00
3 9 0.00
4 9 0.00
5 9 0.00
6 9 0.00
7 9 0.00
8 9 0.00
9 9 1.00
10 9 0.00
11 9 0.00
12 9 0.00
13 9 0.00
14 9 0.00
15 9 0.00
16 9 0.00
17 9 0.00
18 9 0.00
19 9 0.00
20 9 0.00
21 9 0.00
22 9 0.00
23 9 0.00
24 9 0.00
25 9 0.00
0 10 0.00
1 10 0.00
2 10 0.00
3 10 0.00
4 10 0.00
5 10 0.00
6 10 0.00
7 10 0.00
8 10 0.00
9 10 0.00
10 10 0.99
11 10 0.00
12 10 0.00
13 10 0.00
14 10 0.00
15 10 0.00
16 10 0.00
17 10 0.00
18 10 0.00
19 10 0.00
20 10 0.00
21 10 0.00
22 10 0.00
23 10 0.01
24 10 0.00
25 10 0.00
0 11 0.00
1 11 0.00
2 11 0.00
3 11 0.00
4 11 0.00
5 11 0.00
6 11 0.00
7 11 0.00
8 11 0.00
9 11 0.00
10 11 0.00
11 11 1.00
12 11 0.00
13 11 0.00
14 11 0.00
15 11 0.00
16 11 0.00
17 11 0.00
18 11 0.00
19 11 0.00
20 11 0.00
21 11 0.00
22 11 0.00
23 11 0.00
24 11 0.00
25 11 0.00
0 12 0.00
1 12 0.05
2 12 0.00
3 12 0.00
4 12 0.00
5 12 0.00
6 12 0.00
7 12 0.00
8 12 0.00
9 12 0.00
10 12 0.00
11 12 0.00
12 12 0.95
13 12 0.00
14 12 0.00
15 12 0.00
16 12 0.00
17 12 0.00
18 12 0.00
19 12 0.00
20 12 0.00
21 12 0.00
22 12 0.00
23 12 0.00
24 12 0.00
25 12 0.00
0 13 0.00
1 13 0.00
2 13 0.00
3 13 0.00
4 13 0.00
5 13 0.00
6 13 0.00
7 13 0.00
8 13 0.00
9 13 0.00
10 13 0.00
11 13 0.00
12 13 0.03
13 13 0.97
14 13 0.00
15 13 0.00
16 13 0.00
17 13 0.00
18 13 0.00
19 13 0.00
20 13 0.00
21 13 0.00
22 13 0.00
23 13 0.00
24 13 0.00
25 13 0.00
0 14 0.00
1 14 0.00
2 14 0.00
3 14 0.00
4 14 0.00
5 14 0.00
6 14 0.00
7 14 0.00
8 14 0.00
9 14 0.00
10 14 0.00
11 14 0.00
12 14 0.00
13 14 0.00
14 14 1.00
15 14 0.00
16 14 0.00
17 14 0.00
18 14 0.00
19 14 0.00
20 14 0.00
21 14 0.00
22 14 0.00
23 14 0.00
24 14 0.00
25 14 0.00
0 15 0.00
1 15 0.06
2 15 0.00
3 15 0.00
4 15 0.00
5 15 0.00
6 15 0.00
7 15 0.00
8 15 0.00
9 15 0.00
10 15 0.00
11 15 0.00
12 15 0.00
13 15 0.00
14 15 0.00
15 15 0.94
16 15 0.00
17 15 0.00
18 15 0.00
19 15 0.00
20 15 0.00
21 15 0.00
22 15 0.00
23 15 0.00
24 15 0.00
25 15 0.00
0 16 0.00
1 16 0.00
2 16 0.22
3 16 0.00
4 16 0.00
5 16 0.00
6 16 0.00
7 16 0.00
8 16 0.00
9 16 0.00
10 16 0.00
11 16 0.00
12 16 0.00
13 16 0.00
14 16 0.00
15 16 0.00
16 16 0.78
17 16 0.00
18 16 0.00
19 16 0.00
20 16 0.00
21 16 0.00
22 16 0.00
23 16 0.00
24 16 0.00
25 16 0.00
0 17 0.00
1 17 0.00
2 17 0.00
3 17 0.00
4 17 0.00
5 17 0.00
6 17 0.00
7 17 0.00
8 17 0.00
9 17 0.00
10 17 0.00
11 17 0.00
12 17 0.00
13 17 0.00
14 17 0.00
15 17 0.00
16 17 0.00
17 17 1.00
18 17 0.00
19 17 0.00
20 17 0.00
21 17 0.00
22 17 0.00
23 17 0.00
24 17 0.00
25 17 0.00
0 18 0.00
1 18 0.00
2 18 0.00
3 18 0.00
4 18 0.00
5 18 0.00
6 18 0.00
7 18 0.00
8 18 0.00
9 18 0.00
10 18 0.00
11 18 0.00
12 18 0.00
13 18 0.00
14 18 0.00
15 18 0.00
16 18 0.00
17 18 0.00
18 18 0.97
19 18 0.00
20 18 0.00
21 18 0.00
22 18 0.00
23 18 0.00
24 18 0.00
25 18 0.03
0 19 0.00
1 19 0.00
2 19 0.00
3 19 0.00
4 19 0.00
5 19 0.00
6 19 0.00
7 19 0.00
8 19 0.00
9 19 0.00
10 19 0.00
11 19 0.00
12 19 0.00
13 19 0.00
14 19 0.00
15 19 0.00
16 19 0.00
17 19 0.00
18 19 0.00
19 19 1.00
20 19 0.00
21 19 0.00
22 19 0.00
23 19 0.00
24 19 0.00
25 19 0.00
0 20 0.00
1 20 0.00
2 20 0.00
3 20 0.00
4 20 0.00
5 20 0.00
6 20 0.00
7 20 0.00
8 20 0.00
9 20 0.00
10 20 0.00
11 20 0.00
12 20 0.00
13 20 0.00
14 20 0.00
15 20 0.00
16 20 0.00
17 20 0.00
18 20 0.00
19 20 0.00
20 20 0.76
21 20 0.24
22 20 0.00
23 20 0.00
24 20 0.00
25 20 0.00
0 21 0.00
1 21 0.00
2 21 0.00
3 21 0.00
4 21 0.00
5 21 0.00
6 21 0.04
7 21 0.08
8 21 0.00
9 21 0.00
10 21 0.00
11 21 0.00
12 21 0.00
13 21 0.00
14 21 0.00
15 21 0.00
16 21 0.00
17 21 0.00
18 21 0.04
19 21 0.00
20 21 0.23
21 21 0.62
22 21 0.00
23 21 0.00
24 21 0.00
25 21 0.00
0 22 0.00
1 22 0.00
2 22 0.00
3 22 0.00
4 22 0.00
5 22 0.00
6 22 0.00
7 22 0.00
8 22 0.00
9 22 0.00
10 22 0.00
11 22 0.00
12 22 0.00
13 22 0.00
14 22 0.00
15 22 0.00
16 22 0.00
17 22 0.00
18 22 0.00
19 22 0.00
20 22 0.00
21 22 0.00
22 22 1.00
23 22 0.00
24 22 0.00
25 22 0.00
0 23 0.00
1 23 0.00
2 23 0.00
3 23 0.00
4 23 0.00
5 23 0.00
6 23 0.00
7 23 0.00
8 23 0.00
9 23 0.00
10 23 0.00
11 23 0.00
12 23 0.00
13 23 0.00
14 23 0.00
15 23 0.00
16 23 0.00
17 23 0.00
18 23 0.00
19 23 0.00
20 23 0.00
21 23 0.00
22 23 0.00
23 23 1.00
24 23 0.00
25 23 0.00
0 24 0.00
1 24 0.00
2 24 0.00
3 24 0.00
4 24 0.00
5 24 0.00
6 24 0.00
7 24 0.00
8 24 0.00
9 24 0.00
10 24 0.00
11 24 0.00
12 24 0.00
13 24 0.00
14 24 0.00
15 24 0.00
16 24 0.00
17 24 0.00
18 24 0.00
19 24 0.00
20 24 0.00
21 24 0.00
22 24 0.00
23 24 0.00
24 24 1.00
25 24 0.00
0 25 0.00
1 25 0.00
2 25 0.04
3 25 0.00
4 25 0.00
5 25 0.00
6 25 0.00
7 25 0.00
8 25 0.00
9 25 0.00
10 25 0.00
11 25 0.00
12 25 0.00
13 25 0.00
14 25 0.00
15 25 0.00
16 25 0.00
17 25 0.00
18 25 0.00
19 25 0.00
20 25 0.00
21 25 0.00
22 25 0.00
23 25 0.00
24 25 0.00
25 25 0.95
        };
    \end{axis}
\end{tikzpicture}
%

%% file: figures/train3.tex
\begin{tikzpicture}[scale=0.8]
\begin{axis}[axis y line*=left,
		   width=0.45\textwidth,
		   height=0.4\textwidth,
		   /pgf/number format/1000 sep={},
                    xmin=0,xmax=6,
                    ymin=0.0,ymax=0.2,
                    xticklabel style={
                    	/pgf/number format/fixed,
			/pgf/number format/precision=0},
                    yticklabel style={
                    	/pgf/number format/fixed,
			/pgf/number format/precision=2},
                    xlabel={Epoch},
                    ylabel={Loss}] 
\addplot[color=green,ultra thick,mark=star,mark size=2.0] coordinates {
(0,0.129612 )
(1,0.117508 )
(2,0.109505 )
(3,0.100675 )
(4,0.104441 )
(5,0.100405 )
(6,0.092951 )
}; \label{plot_one3}
\addlegendentry{Training loss}
\addplot[color=red,ultra thick,mark=star,mark size=2.0] coordinates {
(0,0.087197)
(1,0.079279)
(2,0.085348)
(3,0.085141)
(4,0.082962)
(5,0.078189)
(6,0.077367)
}; \label{plot_two3}
\addlegendentry{Evaluation loss}
\end{axis}
\begin{axis}[axis y line*=right,
		   width=0.45\textwidth,
		   height=0.4\textwidth,
		   /pgf/number format/1000 sep={},
                    xmin=0,xmax=6,
                    ymin=0.96,ymax=1.0,
                    xticklabel style={
                    	/pgf/number format/fixed,
			/pgf/number format/precision=0},
                    yticklabel style={
                    	/pgf/number format/fixed,
			/pgf/number format/precision=2},
                    xlabel={Epoch},
                    ylabel={Accuracy}] 
\addlegendimage{/pgfplots/refstyle=plot_one3}\addlegendentry{Training loss}
\addlegendimage{/pgfplots/refstyle=plot_two3}\addlegendentry{Evaluation loss}
\addplot[color=blue,ultra thick,mark=star,mark size=2.0] coordinates {
(0,0.973392)
(1,0.976569)
(2,0.974980)
(3,0.975377)
(4,0.975377)
(5,0.976966)
(6,0.976172)
}; \label{plot_three3}
\addlegendentry{Accuracy}
\end{axis}
\end{tikzpicture}

%% file: figures/conf_exp3a.tex
\begin{tikzpicture}[scale=0.8]
    \begin{axis}[
        width=6cm,
        height=6cm,
	colormap={bluewhite}{color=(white) rgb255=(100,149,237)},
        xticklabels={benign,malware},
        xtick={0,...,1},
        xtick style={draw=none},
	xticklabel style={anchor=east,rotate=45},
        yticklabels={benign,malware},
        ytick={0,...,1},
        ytick style={draw=none},
        enlargelimits=false,
        xlabel style={font=\footnotesize},
        ylabel style={font=\footnotesize},
        legend style={font=\footnotesize},
        xticklabel style={font=\footnotesize},
        yticklabel style={font=\footnotesize},
        colorbar,
        colorbar style={
            ytick={0,0.20,0.40,0.60,0.80,1.00},
            yticklabels={0,0.20,0.40,0.60,0.80,1.00},
            yticklabel={\pgfmathprintnumber\tick},
            yticklabel style={font=\footnotesize,
            		/pgf/number format/fixed,
			/pgf/number format/precision=1}
        },
        point meta min=0.0,
        point meta max=1.0,
        nodes near coords={\pgfmathprintnumber\pgfplotspointmeta},
        nodes near coords black white/.style={
            small value/.style={
                font=\footnotesize,
                yshift=-7pt,
                text=black,
                /pgf/number format/fixed,
                /pgf/number format/precision=2
            },
            large value/.style={
                font=\footnotesize,
                yshift=-7pt,
                text=white,
                /pgf/number format/fixed,
                /pgf/number format/precision=2
            },
            every node near coord/.style={
                check for zero/.code={
                    \pgfmathfloatifflags{\pgfplotspointmeta}{0}{
                        \pgfkeys{/tikz/coordinate}
                    }{
                        \begingroup
                        \pgfkeys{/pgf/fpu}
                        \pgfmathparse{\pgfplotspointmeta<#1}
                        \global\let\result=\pgfmathresult
                        \endgroup
                        %
                        %
                        \pgfmathfloatcreate{1}{1.0}{0}
                        \let\ONE=\pgfmathresult
                        \ifx\result\ONE
                            \pgfkeysalso{/pgfplots/small value}
                        \else
                            \pgfkeysalso{/pgfplots/large value}
                        \fi
                    }
                },
                check for zero,
            },
        },
        nodes near coords black white=0.5,
    ]
        \addplot[
            matrix plot,
            mesh/cols=2,
            point meta=explicit,draw=gray
        ] table [meta=C] {
            x y C
            0 0 0.91
            1 0 0.09
            0 1 0.00
            1 1 1.00
        };
    \end{axis}
\end{tikzpicture}
%

%% file: figures/train4.tex
\begin{tikzpicture}[scale=0.8]
\begin{axis}[axis y line*=left,
		   width=0.45\textwidth,
		   height=0.4\textwidth,
		   /pgf/number format/1000 sep={},
                    xmin=0,xmax=6,
                    ymin=0.2,ymax=0.8,
                    xticklabel style={
                    	/pgf/number format/fixed,
			/pgf/number format/precision=0},
                    yticklabel style={
                    	/pgf/number format/fixed,
			/pgf/number format/precision=2},
                    xlabel={Epoch},
                    ylabel={Loss}] 
\addplot[color=green,ultra thick,mark=star,mark size=2.0] coordinates {
(0,0.454778)
(1,0.389076)
(2,0.390634)
(3,0.392197)
(4,0.361965)
(5,0.369034)
(6,0.339458)
}; \label{plot_one4}
\addlegendentry{Training loss}
\addplot[color=red,ultra thick,mark=star,mark size=2.0] coordinates {
(0,0.266330)
(1,0.264421)
(2,0.261393)
(3,0.272218)
(4,0.267454)
(5,0.258368)
(6,0.264583)
}; \label{plot_two4}
\addlegendentry{Evaluation loss}
\end{axis}
\begin{axis}[axis y line*=right,
		   width=0.45\textwidth,
		   height=0.4\textwidth,
		   /pgf/number format/1000 sep={},
                    xmin=0,xmax=6,
                    ymin=0.88,ymax=0.98,
                    xticklabel style={
                    	/pgf/number format/fixed,
			/pgf/number format/precision=0},
                    yticklabel style={
                    	/pgf/number format/fixed,
			/pgf/number format/precision=2},
                    xlabel={Epoch},
                    ylabel={Accuracy}] 
\addlegendimage{/pgfplots/refstyle=plot_one4}\addlegendentry{Training loss}
\addlegendimage{/pgfplots/refstyle=plot_two4}\addlegendentry{Evaluation loss}
\addplot[color=blue,ultra thick,mark=star,mark size=2.0] coordinates {
(0,0.926868)
(1,0.929253)
(2,0.930843)
(3,0.923291)
(4,0.928855)
(5,0.933227)
(6,0.929253)
}; \label{plot_three4}
\addlegendentry{Accuracy}
\end{axis}
\end{tikzpicture}

%% file: figures/conf_exp4.tex
\begin{tikzpicture}[scale=0.8]
    \begin{axis}[
        width=6cm,
        height=6cm,
	colormap={bluewhite}{color=(white) rgb255=(100,149,237)},
        xticklabels={benign,securityshield,winwebsec,zbot,zeroaccess},
        xtick={0,...,4},
        xtick style={draw=none},
	xticklabel style={anchor=east,rotate=45},
        yticklabels={benign,securityshield,winwebsec,zbot,zeroaccess},
        ytick={0,...,4},
        ytick style={draw=none},
        enlargelimits=false,
        xlabel style={font=\footnotesize},
        ylabel style={font=\footnotesize},
        legend style={font=\footnotesize},
        xticklabel style={font=\footnotesize},
        yticklabel style={font=\footnotesize},
        colorbar,
        colorbar style={
            ytick={0,0.20,0.40,0.60,0.80,1.00},
            yticklabels={0,0.20,0.40,0.60,0.80,1.00},
            yticklabel={\pgfmathprintnumber\tick},
            yticklabel style={font=\footnotesize,
            		/pgf/number format/fixed,
			/pgf/number format/precision=1}
        },
        point meta min=0.0,
        point meta max=1.0,
        nodes near coords={\pgfmathprintnumber\pgfplotspointmeta},
        nodes near coords black white/.style={
            small value/.style={
                font=\footnotesize,
                yshift=-7pt,
                text=black,
                /pgf/number format/fixed,
                /pgf/number format/precision=2
            },
            large value/.style={
                font=\footnotesize,
                yshift=-7pt,
                text=white,
                /pgf/number format/fixed,
                /pgf/number format/precision=2
            },
            every node near coord/.style={
                check for zero/.code={
                    \pgfmathfloatifflags{\pgfplotspointmeta}{0}{
                        \pgfkeys{/tikz/coordinate}
                    }{
                        \begingroup
                        \pgfkeys{/pgf/fpu}
                        \pgfmathparse{\pgfplotspointmeta<#1}
                        \global\let\result=\pgfmathresult
                        \endgroup
                        %
                        %
                        \pgfmathfloatcreate{1}{1.0}{0}
                        \let\ONE=\pgfmathresult
                        \ifx\result\ONE
                            \pgfkeysalso{/pgfplots/small value}
                        \else
                            \pgfkeysalso{/pgfplots/large value}
                        \fi
                    }
                },
                check for zero,
            },
        },
        nodes near coords black white=0.5,
    ]
        \addplot[
            matrix plot,
            mesh/cols=5,
            point meta=explicit,draw=gray
        ] table [meta=C] {
            x y C
            0 0 0.91
            1 0 0.00
            2 0 0.06
            3 0 0.02
            4 0 0.02
            0 1 0.00
            1 1 0.00
            2 1 1.00
            3 1 0.00
            4 1 0.00
            0 2 0.00
            1 2 0.00
            2 2 1.00
            3 2 0.00
            4 2 0.00
            0 3 0.01
            1 3 0.00
            2 3 0.10
            3 3 0.89
            4 3 0.00
            0 4 0.00
            1 4 0.00
            2 4 0.14
            3 4 0.02
            4 4 0.84
        };
    \end{axis}
\end{tikzpicture}
%

%% file: figures/optK_5.tex
\begin{tikzpicture}[scale=0.8]
\begin{axis}[
		   width=0.45\textwidth,
		   height=0.4\textwidth,
		   /pgf/number format/1000 sep={},
                    xmin=0.9,xmax=9.1,
                    ymin=0.990,ymax=1.0,
		   xtick={1,2,3,4,5,6,7,8,9},
                    xticklabel style={
                    	/pgf/number format/fixed,
			/pgf/number format/precision=0},
                    yticklabel style={
                    	/pgf/number format/fixed,
			/pgf/number format/precision=3},
                    xlabel={Number of neighbors ($k$)},
                    ylabel={Accuracy}] 
\addplot[color=blue,ultra thick,mark=star,mark size=2.0] coordinates {
(1,0.996)
(2,0.995)
(3,0.9945)
(4,0.994)
(5,0.9937)
(6,0.9941)
(7,0.9935)
(8,0.993)
(9,0.992)
};
\addplot[color=red,ultra thick,mark=star,mark size=2.0] coordinates {
((1,1.0)
(2,0.998)
(3,0.997)
(4,0.9965)
(5,0.9955)
(6,0.9957)
(7,0.9945)
(8,0.9946)
(9,0.992)
};
\legend{Testing,Training}
\end{axis}
\end{tikzpicture}

%% file: figures/conf_exp5a_b.tex
\begin{tikzpicture}[scale=0.8]
    \begin{axis}[
        width=6cm,
        height=6cm,
	colormap={bluewhite}{color=(white) rgb255=(100,149,237)},
        xticklabels={benign,malware},
        xtick={0,...,1},
        xtick style={draw=none},
	xticklabel style={anchor=east,rotate=45},
        yticklabels={benign,malware},
        ytick={0,...,1},
        ytick style={draw=none},
        enlargelimits=false,
        xlabel style={font=\footnotesize},
        ylabel style={font=\footnotesize},
        legend style={font=\footnotesize},
        xticklabel style={font=\footnotesize},
        yticklabel style={font=\footnotesize},
        colorbar,
        colorbar style={
            ytick={0,0.20,0.40,0.60,0.80,1.00},
            yticklabels={0,0.20,0.40,0.60,0.80,1.00},
            yticklabel={\pgfmathprintnumber\tick},
            yticklabel style={font=\footnotesize,
            		/pgf/number format/fixed,
			/pgf/number format/precision=1}
        },
        point meta min=0.0,
        point meta max=1.0,
        nodes near coords={\pgfmathprintnumber\pgfplotspointmeta},
        nodes near coords black white/.style={
            small value/.style={
                font=\footnotesize,
                yshift=-7pt,
                text=black,
                /pgf/number format/fixed,
                /pgf/number format/precision=2
            },
            large value/.style={
                font=\footnotesize,
                yshift=-7pt,
                text=white,
                /pgf/number format/fixed,
                /pgf/number format/precision=2
            },
            every node near coord/.style={
                check for zero/.code={
                    \pgfmathfloatifflags{\pgfplotspointmeta}{0}{
                        \pgfkeys{/tikz/coordinate}
                    }{
                        \begingroup
                        \pgfkeys{/pgf/fpu}
                        \pgfmathparse{\pgfplotspointmeta<#1}
                        \global\let\result=\pgfmathresult
                        \endgroup
                        %
                        %
                        \pgfmathfloatcreate{1}{1.0}{0}
                        \let\ONE=\pgfmathresult
                        \ifx\result\ONE
                            \pgfkeysalso{/pgfplots/small value}
                        \else
                            \pgfkeysalso{/pgfplots/large value}
                        \fi
                    }
                },
                check for zero,
            },
        },
        nodes near coords black white=0.5,
    ]
        \addplot[
            matrix plot,
            mesh/cols=2,
            point meta=explicit,draw=gray
        ] table [meta=C] {
            x y C
            0 0 0.99
            1 0 0.01
            0 1 0.00
            1 1 1.00
        };
    \end{axis}
\end{tikzpicture}
%

%% file: figures/optK_6.tex
\begin{tikzpicture}[scale=0.8]
\begin{axis}[
		   width=0.45\textwidth,
		   height=0.4\textwidth,
		   /pgf/number format/1000 sep={},
                    xmin=0.9,xmax=9.1,
                    ymin=0.984,ymax=1.0,
		   xtick={1,2,3,4,5,6,7,8,9},
		   ytick={0.984,0.988,0.992,0.996,1.000},
                    xticklabel style={
                    	/pgf/number format/fixed,
			/pgf/number format/precision=0},
                    yticklabel style={
                    	/pgf/number format/fixed,
			/pgf/number format/precision=3},
                    xlabel={Number of neighbors ($k$)},
                    ylabel={Accuracy}] 
\addplot[color=blue,ultra thick,mark=star,mark size=2.0] coordinates {
(1,0.994)
(2,0.9932)
(3,0.9925)
(4,0.9915)
(5,0.990)
(6,0.9902)
(7,0.989)
(8,0.9885)
(9,0.986)
};
\addplot[color=red,ultra thick,mark=star,mark size=2.0] coordinates {
((1,1.0)
(2,0.9985)
(3,0.9973)
(4,0.9966)
(5,0.9945)
(6,0.9945)
(7,0.994)
(8,0.993)
(9,0.992)
};
\legend{Testing,Training}
\end{axis}
\end{tikzpicture}

%% file: figures/conf_exp6.tex
\begin{tikzpicture}[scale=0.8]
    \begin{axis}[
        width=6cm,
        height=6cm,
	colormap={bluewhite}{color=(white) rgb255=(100,149,237)},
        xticklabels={benign,securityshield,zeroaccess,zbot,winwebsec},
        xtick={0,...,4},
        xtick style={draw=none},
	xticklabel style={anchor=east,rotate=45},
        yticklabels={benign,securityshield,zeroaccess,zbot,winwebsec},
        ytick={0,...,4},
        ytick style={draw=none},
        enlargelimits=false,
        xlabel style={font=\footnotesize},
        ylabel style={font=\footnotesize},
        legend style={font=\footnotesize},
        xticklabel style={font=\footnotesize},
        yticklabel style={font=\footnotesize},
        colorbar,
        colorbar style={
            ytick={0,0.20,0.40,0.60,0.80,1.00},
            yticklabels={0,0.20,0.40,0.60,0.80,1.00},
            yticklabel={\pgfmathprintnumber\tick},
            yticklabel style={font=\footnotesize,
            		/pgf/number format/fixed,
			/pgf/number format/precision=1}
        },
        point meta min=0.0,
        point meta max=1.0,
        nodes near coords={\pgfmathprintnumber\pgfplotspointmeta},
        nodes near coords black white/.style={
            small value/.style={
                font=\footnotesize,
                yshift=-7pt,
                text=black,
                /pgf/number format/fixed,
                /pgf/number format/precision=2
            },
            large value/.style={
                font=\footnotesize,
                yshift=-7pt,
                text=white,
                /pgf/number format/fixed,
                /pgf/number format/precision=2
            },
            every node near coord/.style={
                check for zero/.code={
                    \pgfmathfloatifflags{\pgfplotspointmeta}{0}{
                        \pgfkeys{/tikz/coordinate}
                    }{
                        \begingroup
                        \pgfkeys{/pgf/fpu}
                        \pgfmathparse{\pgfplotspointmeta<#1}
                        \global\let\result=\pgfmathresult
                        \endgroup
                        %
                        %
                        \pgfmathfloatcreate{1}{1.0}{0}
                        \let\ONE=\pgfmathresult
                        \ifx\result\ONE
                            \pgfkeysalso{/pgfplots/small value}
                        \else
                            \pgfkeysalso{/pgfplots/large value}
                        \fi
                    }
                },
                check for zero,
            },
        },
        nodes near coords black white=0.5,
    ]
        \addplot[
            matrix plot,
            mesh/cols=5,
            point meta=explicit,draw=gray
        ] table [meta=C] {
            x y C
            0 0 0.99
            1 0 0.01
            2 0 0.00
            3 0 0.00
            4 0 0.00
            0 1 0.00
            1 1 1.00
            2 1 0.00
            3 1 0.00
            4 1 0.00
            0 2 0.00
            1 2 0.00
            2 2 0.99
            3 2 0.01
            4 2 0.00
            0 3 0.01
            1 3 0.00
            2 3 0.01
            3 3 0.98
            4 3 0.00
            0 4 0.00
            1 4 0.00
            2 4 0.00
            3 4 0.00
            4 4 1.00
        };
    \end{axis}
\end{tikzpicture}
%

%% file: figures/conf_zd.tex
\begin{tabular}{ccc}
\begin{tikzpicture}[scale=0.775]
    \begin{axis}[
        width=6cm,
        height=6cm,
	colormap={bluewhite}{color=(white) rgb255=(100,149,237)},
        xticklabels={benign,malware},
        xtick={0,...,1},
        xtick style={draw=none},
	xticklabel style={anchor=east,rotate=45},
        yticklabels={benign,malware},
        ytick={0,...,1},
        ytick style={draw=none},
        enlargelimits=false,
        xlabel style={font=\footnotesize},
        ylabel style={font=\footnotesize},
        legend style={font=\footnotesize},
        xticklabel style={font=\footnotesize},
        yticklabel style={font=\footnotesize},
        colorbar,
        colorbar style={
            ytick={0,0.20,0.40,0.60,0.80,1.00},
            yticklabels={0,0.20,0.40,0.60,0.80,1.00},
            yticklabel={\pgfmathprintnumber\tick},
            yticklabel style={font=\footnotesize,
            		/pgf/number format/fixed,
			/pgf/number format/precision=1}
        },
        point meta min=0.0,
        point meta max=1.0,
        nodes near coords={\pgfmathprintnumber\pgfplotspointmeta},
        nodes near coords black white/.style={
            small value/.style={
                font=\footnotesize,
                yshift=-7pt,
                text=black,
                /pgf/number format/fixed,
                /pgf/number format/precision=2
            },
            large value/.style={
                font=\footnotesize,
                yshift=-7pt,
                text=white,
                /pgf/number format/fixed,
                /pgf/number format/precision=2
            },
            every node near coord/.style={
                check for zero/.code={
                    \pgfmathfloatifflags{\pgfplotspointmeta}{0}{
                        \pgfkeys{/tikz/coordinate}
                    }{
                        \begingroup
                        \pgfkeys{/pgf/fpu}
                        \pgfmathparse{\pgfplotspointmeta<#1}
                        \global\let\result=\pgfmathresult
                        \endgroup
                        %
                        %
                        \pgfmathfloatcreate{1}{1.0}{0}
                        \let\ONE=\pgfmathresult
                        \ifx\result\ONE
                            \pgfkeysalso{/pgfplots/small value}
                        \else
                            \pgfkeysalso{/pgfplots/large value}
                        \fi
                    }
                },
                check for zero,
            },
        },
        nodes near coords black white=0.5,
    ]
        \addplot[
            matrix plot,
            mesh/cols=2,
            point meta=explicit,draw=gray
        ] table [meta=C] {
            x y C
            0 0 0.99
            1 0 0.01
            0 1 0.21
            1 1 0.79
        };
    \end{axis}
\end{tikzpicture}
&
&
\begin{tikzpicture}[scale=0.775]
    \begin{axis}[
        width=6cm,
        height=6cm,
	colormap={bluewhite}{color=(white) rgb255=(100,149,237)},
        xticklabels={benign,malware},
        xtick={0,...,1},
        xtick style={draw=none},
	xticklabel style={anchor=east,rotate=45},
        yticklabels={benign,malware},
        ytick={0,...,1},
        ytick style={draw=none},
        enlargelimits=false,
        xlabel style={font=\footnotesize},
        ylabel style={font=\footnotesize},
        legend style={font=\footnotesize},
        xticklabel style={font=\footnotesize},
        yticklabel style={font=\footnotesize},
        colorbar,
        colorbar style={
            ytick={0,0.20,0.40,0.60,0.80,1.00},
            yticklabels={0,0.20,0.40,0.60,0.80,1.00},
            yticklabel={\pgfmathprintnumber\tick},
            yticklabel style={font=\footnotesize,
            		/pgf/number format/fixed,
			/pgf/number format/precision=1}
        },
        point meta min=0.0,
        point meta max=1.0,
        nodes near coords={\pgfmathprintnumber\pgfplotspointmeta},
        nodes near coords black white/.style={
            small value/.style={
                font=\footnotesize,
                yshift=-7pt,
                text=black,
                /pgf/number format/fixed,
                /pgf/number format/precision=2
            },
            large value/.style={
                font=\footnotesize,
                yshift=-7pt,
                text=white,
                /pgf/number format/fixed,
                /pgf/number format/precision=2
            },
            every node near coord/.style={
                check for zero/.code={
                    \pgfmathfloatifflags{\pgfplotspointmeta}{0}{
                        \pgfkeys{/tikz/coordinate}
                    }{
                        \begingroup
                        \pgfkeys{/pgf/fpu}
                        \pgfmathparse{\pgfplotspointmeta<#1}
                        \global\let\result=\pgfmathresult
                        \endgroup
                        %
                        %
                        \pgfmathfloatcreate{1}{1.0}{0}
                        \let\ONE=\pgfmathresult
                        \ifx\result\ONE
                            \pgfkeysalso{/pgfplots/small value}
                        \else
                            \pgfkeysalso{/pgfplots/large value}
                        \fi
                    }
                },
                check for zero,
            },
        },
        nodes near coords black white=0.5,
    ]
        \addplot[
            matrix plot,
            mesh/cols=2,
            point meta=explicit,draw=gray
        ] table [meta=C] {
            x y C
            0 0 1.00
            1 0 0.00
            0 1 0.28
            1 1 0.72
        };
    \end{axis}
\end{tikzpicture}
\\
(a) DL results
&
&
(b) $k$-NN results
\\[2ex]
\end{tabular}
%

%% file: ms.bbl
\begin{thebibliography}{}

\bibitem[Austin et~al., 2013]{AustinFJS13}
Austin, T.~H., Filiol, E., Josse, S., and Stamp, M. (2013).
\newblock Exploring hidden {M}arkov models for virus analysis: {A} semantic
  approach.
\newblock In {\em 46th Hawaii International Conference on System Sciences,
  {HICSS} 2013, Wailea, HI, USA, January 7-10, 2013}, pages 5039--5048. {IEEE}
  Computer Society.

\bibitem[Baysa et~al., 2013]{BaysaLS13}
Baysa, D., Low, R.~M., and Stamp, M. (2013).
\newblock Structural entropy and metamorphic malware.
\newblock {\em Journal of Computer Virology and Hacking Techniques},
  9(4):179--192.

\bibitem[Damodaran et~al., 2017]{DamodaranTVAS17}
Damodaran, A., Troia, F.~D., Visaggio, C.~A., Austin, T.~H., and Stamp, M.
  (2017).
\newblock A comparison of static, dynamic, and hybrid analysis for malware
  detection.
\newblock {\em Journal of Computer Virology and Hacking Techniques},
  13(1):1--12.

\bibitem[Fast.ai, 2018]{PB10}
Fast.ai (2018).
\newblock Fast.ai lectures.
\newblock \url{https://course.fast.ai/lessons/lessons.html}.

\bibitem[He et~al., 2016]{PB9}
He, K., Zhang, X., Ren, S., and Sun, J. (2016).
\newblock Deep residual learning for image recognition.
\newblock In {\em 2016 IEEE Conference on Computer Vision and Pattern
  Recognition}, CVPR 2016, pages 770--778.

\bibitem[Machine Learning, 2018]{PB5}
Machine Learning (2018).
\newblock Machine learning: Github repository.
\newblock
  \url{https://github.com/tuff96/Malware-detection-using-Machine-Learning}.

\bibitem[Nappa et~al., 2015]{PB3}
Nappa, A., Rafique, M.~Z., and Caballero, J. (2015).
\newblock The {M}alicia dataset: Identification and analysis of drive-by
  download operations.
\newblock {\em International Journal of Information Security}, 14(1):15--33.

\bibitem[Nataraj et~al., 2011]{PB1}
Nataraj, L., Karthikeyan, S., Jacob, G., and Manjunath, B. (2011).
\newblock Malware images: Visualization and automatic classification.
\newblock In {\em Proceedings of the 8th International Symposium on
  Visualization for Cyber Security}, VizSec '11.

\bibitem[PE File, 2018]{PB4}
PE File (2018).
\newblock Pe file: Github repository.
\newblock \url{https://github.com/erocarrera/pefile}.

\bibitem[Pedregosa et~al., 2011]{PB7}
Pedregosa, F., Varoquaux, G., Gramfort, A., Michel, V., Thirion, B., Grisel,
  O., Blondel, M., Prettenhofer, P., Weiss, R., Dubourg, V., Vanderplas, J.,
  Passos, A., Cournapeau, D., Brucher, M., Perrot, M., and Duchesnay, E.
  (2011).
\newblock Scikit-learn: Machine learning in python.
\newblock {\em J. Mach. Learn. Res.}, 12:2825--2830.

\bibitem[Singh et~al., 2016]{SinghTVAS16}
Singh, T., Troia, F.~D., Visaggio, C.~A., Austin, T.~H., and Stamp, M. (2016).
\newblock Support vector machines and malware detection.
\newblock {\em Journal of Computer Virology and Hacking Techniques},
  12(4):203--212.

\bibitem[Smith, 2015]{PB11}
Smith, L.~N. (2015).
\newblock Cyclical learning rates for training neural networks.
\newblock \url{https://arxiv.org/abs/1506.01186}.

\bibitem[Stamp, 2017]{StampML2017}
Stamp, M. (2017).
\newblock {\em Introduction to Machine Learning with Applications in
  Information Security}.
\newblock Chapman and Hall/CRC, Boca Raton.

\bibitem[Toderici and Stamp, 2013]{TodericiS13}
Toderici, A.~H. and Stamp, M. (2013).
\newblock Chi-squared distance and metamorphic virus detection.
\newblock {\em Journal of Computer Virology and Hacking Techniques},
  9(1):1--14.

\bibitem[Wong and Stamp, 2006]{WongS06}
Wong, W. and Stamp, M. (2006).
\newblock Hunting for metamorphic engines.
\newblock {\em Journal in Computer Virology}, 2(3):211--229.

\bibitem[Yajamanam et~al., 2018]{PB2}
Yajamanam, S., Selvin, V. R.~S., Troia, F.~D., and Stamp, M. (2018).
\newblock Deep learning versus gist descriptors for image-based malware
  classification.
\newblock In {\em Proceedings of the 4th International Conference on
  Information Systems Security and Privacy}, ICISSP 2018, pages 553--561.

\end{thebibliography}
